\documentclass{article}

\usepackage{microtype}
\usepackage{graphicx}
\usepackage{subcaption}
\usepackage{booktabs}

\usepackage{hyperref}

\usepackage[accepted]{icml2026}

\makeatletter
\renewcommand{\ICML@appearing}{\textit{Accepted at the ICML 2026 Workshop on
Resource-Adaptive Foundation Model Inference (AdaptFM), Seoul, South Korea.}
Copyright 2026 by the author(s).}
\makeatother

\usepackage{amsmath}
\usepackage{amssymb}
\usepackage{mathtools}
\usepackage{amsthm}
\usepackage{multirow}
\usepackage{array}
\usepackage{xcolor}
\usepackage{xspace}
\usepackage{enumitem}

\usepackage[capitalize,noabbrev]{cleveref}

\theoremstyle{plain}

\theoremstyle{definition}

\theoremstyle{remark}

\newcommand{\method}{\textsc{LEAP}\xspace}

\icmltitlerunning{LEAP: Learnable End-to-End Adaptive Pruning of LLMs}

\begin{document}

\twocolumn[
  \icmltitle{\method: Learnable End-to-End Adaptive Pruning \\
    of Large Language Models}

  \icmlsetsymbol{equal}{*}

  \begin{icmlauthorlist}
    \icmlauthor{Mohammad Mozaffari}{elastix,uoft}
    \icmlauthor{Younes Hourri}{uoft}
    \icmlauthor{Mohammad Rastegari}{elastix}
    \icmlauthor{Mahyar Najibi}{elastix}
  \end{icmlauthorlist}

  \icmlaffiliation{elastix}{Elastix AI}
  \icmlaffiliation{uoft}{Department of Computer Science, University of Toronto, Toronto, Ontario, Canada}

  \icmlcorrespondingauthor{Mohammad Mozaffari}{mmozaffari@elastix.ai}

  \icmlkeywords{LLM pruning, unstructured sparsity, learnable masks, model compression}

  \vskip 0.3in
]

\printAffiliationsAndNotice{}  

\begin{abstract}
Unstructured sparsity is now natively accelerated by recent GPU kernels
and dataflow hardware, shifting the bottleneck from inference execution
to the pruning algorithm. State-of-the-art methods for unstructured LLM
pruning are layer-wise surrogates derived from the Optimal Brain Surgeon
principle, and they sacrifice end-to-end accuracy, especially under
aggressive sparsity. End-to-end alternatives such as MaskLLM and PATCH
show that learnable masks can close this gap, but their
categorical-over-patterns parameterization scales with the number of
valid masks per row and does not port to the unstructured setting. We
introduce \method, which replaces this intractable parameterization with
a per-weight Bernoulli-via-Gumbel-Sigmoid relaxation that makes
end-to-end unstructured mask learning tractable. Across five LLM
families from $0.5$B to $8$B parameters at $50\%$ and $60\%$ sparsity,
\method{} improves six-task average zero-shot accuracy by $+2.59$
points on average over ADMM, the best layer-wise baseline in our
sweep.\footnote{Code is available at
\url{https://github.com/Paramathic/patch/tree/leap}.}

\end{abstract}

\section{Introduction}
\label{sec:intro}

Deployment of modern Large Language Models (LLMs) is bottlenecked by memory
and compute, and weight sparsity has become a central tool for reducing
both. Sparsity patterns split into three regimes: structured, semi-structured
(\eg, 2:4), and unstructured. Structured and semi-structured variants enjoy
native GPU support but trade non-trivial accuracy for modest compression.
Unstructured sparsity retains far higher accuracy, and recent kernel work
(SpInfer~\cite{spinfer}, FlashLLM~\cite{flashllm},
MACKO~\cite{macko}) together with sparsity-native dataflow
hardware~\cite{cerebras} now converts unstructured masks into real
speedups on commodity GPUs and wafer-scale engines. The bottleneck has
therefore shifted: the open problem is no longer how to execute unstructured
sparsity but how to \emph{induce} it with minimal accuracy loss.

The dominant algorithmic family for unstructured LLM pruning follows the
Optimal Brain Surgeon (OBS)~\cite{obs} lineage. Wanda~\cite{wanda},
SparseGPT~\cite{sparsegpt}, Thanos~\cite{thanos}, ADMM~\cite{admm}, and
OPTIMA~\cite{optima} all minimize a \emph{layer-wise} reconstruction
error as a surrogate for end-to-end model loss. This surrogate is cheap but misaligned with the
quantity actually being optimized, and it accumulates local errors that
compound in deep networks. Learnable-mask methods such as
MaskLLM~\cite{maskllm} and PATCH~\cite{patch} instead directly optimize masks
with respect to the language modeling loss. They deliver state-of-the-art
results but only for semi-structured patterns.

MaskLLM's parameterization assigns one learnable logit to each valid pattern
inside a group and applies a Gumbel-softmax over this set. For 2:4 sparsity,
the number of valid patterns per group of size $4$ is $\binom{4}{2}=6$, which
is tractable. If one attempts to port this categorical-over-patterns scheme
to unstructured $50\%$ sparsity on a row of width $d{=}4096$, the number of
valid masks is $\binom{4096}{2048} \;\approx\; 10^{1229}$,
which cannot be stored, let alone indexed, as a set of logits. The
parameterization that underlies MaskLLM and PATCH therefore \emph{does not
extend} to the unstructured regime, regardless of compute budget. This is
not an engineering inconvenience but a combinatorial obstruction.

\method{} resolves this obstruction by replacing the categorical distribution
over patterns with a product of independent Bernoullis, one per weight, each
relaxed via the Gumbel-sigmoid trick. The parameter count per weight matrix
scales as $O(mn)$ rather than $O(|\{\text{valid patterns}\}|)$, matching the
weight count. This is the natural reformulation that preserves
end-to-end differentiability for unstructured masks at LLM scale; other
per-weight relaxations (\eg, $L_0$ regularization~\cite{l0reg},
continuous sparsification~\cite{contsparse}) are conceptually related and
similarly tractable. A small set of ingredients (Wanda-based initialization, a scale and
temperature schedule, a global sparsity regularizer, and a magnitude-aware
term) stabilize the resulting optimization. We keep pretrained weights
\emph{frozen} as a deliberate scope choice: decoupling mask learning from
weight updates preserves calibration and simplifies deployment, while
remaining compatible with any subsequent fine-tuning or distillation stage.

Our contributions are as follows:
\begin{itemize}[leftmargin=*,itemsep=0pt,topsep=2pt]
\item We identify a combinatorial obstruction that prevents
MaskLLM/PATCH-style categorical-over-patterns parameterizations from being
used for unstructured sparsity, and we propose a per-weight
Bernoulli-via-Gumbel-sigmoid reformulation that makes end-to-end learning
tractable.
\item We present \method, a lightweight end-to-end unstructured pruning
framework that operates on frozen pretrained weights and trains a per-weight
mask in roughly $2{,}000$ iterations on a small general-text calibration
stream.
\item On Qwen-2.5 0.5B, Gemma-3 1B, LLaMA-3.2 1B, LLaMA-3.2 3B, and
LLaMA-3.1 8B at $50\%$ and $60\%$ sparsity, \method{} improves the
six-task average zero-shot accuracy over ADMM, the best layer-wise
baseline in our sweep, by $+2.59$ points on average across the ten
(model, sparsity) settings and by up to $+5.40$ points on LLaMA-3.2 1B
at $60\%$ sparsity.
\end{itemize}

\section{Related Work}
\label{sec:related}

\paragraph{Hardware support for unstructured sparsity.}
Recent kernel work (FlashLLM~\cite{flashllm}, SpInfer~\cite{spinfer},
MACKO~\cite{macko}) achieves significant speedups for unstructured LLM
sparsity at $50\%$--$60\%$ densities on commodity tensor cores, and
wafer-scale dataflow accelerators~\cite{cerebras} target unstructured
patterns natively. These developments make unstructured masks a deployable
compression target rather than a theoretical one.

\paragraph{Layer-wise OBS-derived pruning.}
Wanda~\cite{wanda} prunes based on the product of weight magnitude and
input activation norms. SparseGPT~\cite{sparsegpt} solves a layer-wise
Hessian-based reconstruction problem and jointly updates surviving weights.
Thanos~\cite{thanos} refines this reconstruction with multi-column updates,
ADMM~\cite{admm} alternates between mask and weight updates, and
OPTIMA~\cite{optima} casts the same reconstruction as a quadratic
program. SLiM~\cite{slim} extends the one-shot reconstruction setting to
joint sparse-plus-low-rank-plus-quantized approximations. These methods
all optimize local surrogates; their errors are provably aligned with
global loss only under strong assumptions that do not hold in LLMs.
Concurrently, ELSA~\cite{elsa} dispenses with the layer-wise surrogate
altogether and uses a surrogate-free ADMM formulation to push unstructured
sparsity into extreme regimes ($\sim$$90\%$); we instead target the
$50\%$--$60\%$ range that current accelerated kernels support, so these
higher sparsity ratios are outside the scope of our work.

\paragraph{Per-weight learnable masks in general pruning.}
Per-weight learnable mask parameterizations have been explored in general
neural network pruning, most notably through $L_0$
regularization~\cite{l0reg} and continuous sparsification~\cite{contsparse},
both of which use continuous relaxations of per-weight binary gates.
\method{} differs in three practical ways: (i) we target LLM-scale
unstructured pruning specifically, where prior end-to-end methods
(MaskLLM, PATCH) adopted categorical-over-patterns parameterizations that
do not scale to the unstructured regime; (ii) we initialize from a
one-shot Wanda mask rather than cold-starting, which sharply reduces the
number of training steps required; (iii) we combine a global sparsity
regularizer with magnitude-aware stabilization tailored to frozen
pretrained weights.

\paragraph{End-to-end learnable masks.}
MaskLLM~\cite{maskllm} and PATCH~\cite{patch} optimize masks directly
against the language modeling loss by parameterizing a categorical
distribution over the valid patterns inside each structured group. PATCH
extends this to tile-level hybrids. A separate line of work folds
sparsity into pretraining itself, e.g., SLoPe~\cite{slope}, which
combines double-pruned sparse weights with lazily attached low-rank
adapters. These training-time approaches are complementary to the
post-training mask-learning regime we study. Both
MaskLLM and PATCH are restricted to
semi-structured regimes because their logit table scales with
$|\{\text{valid patterns}\}|$, which is bounded only when the group is
small. For unstructured $50\%$ sparsity over a single row of width $4096$,
this set has cardinality $\binom{4096}{2048}$, making the parameterization
impossible. \method{} is, to our knowledge, the first practical
reformulation that transfers end-to-end mask learning to the unstructured
setting.

\section{\method{}: Method}
\label{sec:method}

\paragraph{Why Per-Pattern Logits Do Not Scale.}
\label{sec:method:obstruction}
Fix a weight matrix $W \in \mathbb{R}^{m \times n}$ and a target density
budget. A categorical-over-patterns parameterization, as used by
MaskLLM and PATCH, partitions each row into groups of size $g$ and stores
a logit vector of length $|\mathcal{P}_g|$ per group, where $\mathcal{P}_g$
is the set of masks satisfying the pattern constraint. For 2:4 sparsity,
$|\mathcal{P}_4|=\binom{4}{2}=6$. For unstructured $\rho$-sparsity over an
entire row of width $n$, the only natural group is the row itself and
$|\mathcal{P}_n|=\binom{n}{\rho n}$. At $n{=}4096$ and $\rho{=}0.5$ this
is $\binom{4096}{2048} \approx 10^{1229}$, which cannot be represented as
a logit table under any storage or indexing scheme. Smaller groups
reintroduce a structural constraint that is exactly what unstructured
sparsity is defined to avoid. We conclude that the per-pattern
parameterization does not admit an unstructured extension.

\method{} replaces the categorical parameterization with a product of
independent Bernoullis, one per weight. For each weight matrix $W \in
\mathbb{R}^{m \times n}$ we introduce a parallel logit matrix $P \in
\mathbb{R}^{m \times n}$ and a stochastic mask
\begin{equation}
M \;=\; \sigma\!\left(\frac{\alpha P + g}{\tau}\right),
\label{eq:mask}
\end{equation}
where $g = -\log(-\log(u))$ is Gumbel noise with
$u \sim \mathrm{Uniform}(0,1)$, $\sigma$ is the sigmoid function,
$\alpha$ is a scale factor, and $\tau$ is a temperature.
\Cref{eq:mask} is the Gumbel-sigmoid relaxation
of a Bernoulli with logit $\alpha P_{ij}$ and scale $\tau$. The parameter
count per weight matrix is exactly $mn$, which is linear in the weight
count and independent of $\rho$. The effective pruned weight is
\begin{equation}
\widetilde{W} \;=\; M \odot W.
\label{eq:wtilde}
\end{equation}
We use soft masks throughout optimization. Hard sampling with
straight-through estimators is unstable at LLM scale, and soft masks keep
gradients well conditioned while the $\alpha,\tau$ schedules below drive
$M$ toward $\{0,1\}$.

We initialize $P$ from a one-shot Wanda~\cite{wanda} mask. Entries selected
by Wanda are set to $+s$ and the rest to $-s$, where $s>0$ is the initial
mask strength. This gives the sigmoid relaxation a reasonable starting
loss and makes the search local rather than cold.

Two lightweight schedules anneal \Cref{eq:mask} from exploratory to
decisive. The scale $\alpha$ is ramped from $\alpha_0$ to $\alpha_T$
(e.g., $25 \to 350$), which amplifies $P$ and pushes $\sigma$ toward $\{0,1\}$.
The temperature $\tau$ is decayed from $\tau_0$ to $\tau_T$
(e.g., $4.0 \to 0.05$), sharpening the sigmoid. Early iterations explore
many candidate supports; later iterations commit.

Let $\rho$ be the target density (e.g., $\rho=0.5$). Let $\widetilde{M}_i$
denote the soft mask for layer $i$ and let $N_i$ denote the number of
parameters in $W_i$ (so $N = \sum_i N_i$ is the total parameter count
over prunable layers). \method{} enforces density \emph{globally}, not
per layer:
\begin{equation}
\mathcal{L}_{\mathrm{sparsity}}
\;=\;
\lambda_1 \left|\,
\frac{1}{N}\sum_i \|\widetilde{M}_i\|_1 - \rho
\,\right|,
\label{eq:sparsity}
\end{equation}
where $\lambda_1$ is a large positive coefficient. The global form lets
individual layers adjust their density based on end-to-end importance.

To bias the optimization toward retaining higher-magnitude weights we
add
\begin{equation}
\mathcal{L}_{\mathrm{weight}}
\;=\; -\,\lambda_2 \sum_i \|\widetilde{W}_i\|_1,
\label{eq:weight}
\end{equation}
with $\lambda_2 > 0$ (typically $\sim$$10$). This term stabilizes mask
learning and avoids degenerate minima that keep many small weights while
dropping a few critical ones.

\paragraph{Full Objective and Scope.}
\label{sec:method:obj}
Combining \Cref{eq:sparsity,eq:weight} with the language modeling loss on
a calibration stream $X$ yields
\begin{equation}
\mathcal{L} \;=\; \mathcal{L}_{\mathrm{LM}}(\widetilde{W}; X)
\;+\; \mathcal{L}_{\mathrm{sparsity}}
\;+\; \mathcal{L}_{\mathrm{weight}}.
\label{eq:objective}
\end{equation}
Only $P$ is trained; $W$ is held fixed. We treat this as a deliberate
scope choice, not a compute concession: freezing $W$ preserves the
calibration of the pretrained weights, isolates the mask as the object
being learned, and keeps the deployment pipeline simple. Joint
weight-and-mask optimization is a natural extension (\Cref{sec:discussion}).

\section{Experiments}
\label{sec:experiments}

\paragraph{Models.}
We evaluate \method{} across Qwen-2.5 0.5B~\cite{qwen25},
Gemma-3 1B~\cite{gemma3}, LLaMA-3.2 1B, LLaMA-3.2 3B~\cite{llama32}, and
LLaMA-3.1 8B~\cite{llama32} at $50\%$ and $60\%$ unstructured sparsity.

\paragraph{Training setup.}
Following the dataset configuration of MaskLLM~\cite{maskllm} and
PATCH~\cite{patch}, masks are trained for $2{,}000$ steps with batch
size $256$ on sequences of length $4096$ from SlimPajama~\cite{slimpajama}.
Weights are frozen. Hyperparameters are in \Cref{app:hparams}.

\paragraph{Evaluation.}
We report WikiText2 perplexity~\cite{wikitext2} at sequence length $4096$
and zero-shot accuracy on six standard benchmarks: PIQA~\cite{piqa},
ARC-Easy and ARC-Challenge~\cite{arc}, Winogrande~\cite{winogrande},
OpenBookQA~\cite{obqa}, and MMLU~\cite{mmlu}, using the
\texttt{lm-evaluation-harness}~\cite{lmeval}.

\paragraph{Baselines.}
We compare against Wanda~\cite{wanda}, SparseGPT~\cite{sparsegpt},
Thanos~\cite{thanos}, and ADMM~\cite{admm}, each using its default
configuration with $128$ C4~\cite{c4} calibration samples.

\paragraph{Main results.}
\Cref{tab:summary} summarizes WikiText2 perplexity and the six-task
zero-shot average accuracy for all five models at $50\%$ and $60\%$
sparsity. Per-task breakdowns for the $0.5$B--$3$B models are deferred to
\Cref{app:per-task}. \method{} consistently outperforms all baselines,
including ADMM, the best layer-wise baseline in our sweep. Averaging
across the ten (model, sparsity) settings, \method{} improves the
six-task average zero-shot accuracy over ADMM by $+2.59$ points, with
the smallest gain being $+0.21$ points (LLaMA-3.1 8B at $50\%$,
$57.71$ vs.\ $57.50$) and the largest being $+5.40$ points (LLaMA-3.2
1B at $60\%$, $50.39$ vs.\ $44.99$). On LLaMA-3.1 8B, \method{} reaches
$7.66$ PPL and $57.71$ average accuracy at $50\%$, and $8.82$ PPL and
$54.47$ average accuracy at $60\%$, against ADMM's $9.12$/$57.50$ and
$14.10$/$50.61$ respectively. Gains widen at higher sparsity: ADMM is
nearly competitive with \method{} on $50\%$ LLaMA-3.1 8B but the
$60\%$ margin reopens to $+3.86$ points.

We include the learnable-mask baseline MaskLLM~\cite{maskllm} at $50\%$.
For Qwen-2.5 0.5B, Gemma-3 1B, and LLaMA-3.2 1B we use the numbers
reported by PATCH~\cite{patch}; for LLaMA-3.1 8B we instead reproduce
MaskLLM ourselves using its publicly released checkpoint, obtaining
$9.17$ WikiText2 perplexity and $55.09$ six-task average accuracy.
Because MaskLLM's categorical-over-patterns parameterization is
restricted to $2{:}4$ (see \Cref{sec:method:obstruction}), its row in
\Cref{tab:summary} is $2{:}4$ semi-structured at $50\%$ density, not
$50\%$ unstructured. We compute its six-task average over the tasks
shared with our evaluation (MMLU, PIQA, ARC-E, ARC-C, Winogrande, OBQA);
RACE and HellaSwag are excluded. LLaMA-3.2 3B is not reported in the
PATCH paper. Note that MaskLLM's small-model $2{:}4$ accuracy
in~\cite{patch} is notably below its large-model results, consistent
with the known difficulty of training $2{:}4$ masks on small backbones.

\begin{table*}[t]
\caption{Summary of pruning results. We report WikiText2 perplexity
(PPL $\downarrow$, sequence length $4096$) and six-task zero-shot average
accuracy (Avg.\ $\uparrow$, averaged over MMLU, PIQA, ARC-E, ARC-C,
Winogrande, OBQA). Per-task breakdowns for the $0.5$B--$3$B models are in
\Cref{app:per-task}. Bold marks the best sparse method in each column of
each sparsity block. MaskLLM$^{\dagger}$ is $2{:}4$ semi-structured (not
unstructured); LLaMA-3.2 3B is not reported in the PATCH paper. The
LLaMA-3.1 8B dense row is computed from~\cite{patch} by averaging over
the six tasks we report (so it differs from PATCH's eight-task averages,
which additionally include RACE and HellaSwag).}
\label{tab:summary}
\centering
\small
\setlength{\tabcolsep}{3pt}
\begin{tabular}{llcccccccccc}
\toprule
 &  & \multicolumn{2}{c}{Qwen-2.5 0.5B} & \multicolumn{2}{c}{Gemma-3 1B} & \multicolumn{2}{c}{LLaMA-3.2 1B} & \multicolumn{2}{c}{LLaMA-3.2 3B} & \multicolumn{2}{c}{LLaMA-3.1 8B} \\
\cmidrule(lr){3-4}\cmidrule(lr){5-6}\cmidrule(lr){7-8}\cmidrule(lr){9-10}\cmidrule(lr){11-12}
Sparsity & Method & PPL$\downarrow$ & Avg.$\uparrow$ & PPL$\downarrow$ & Avg.$\uparrow$ & PPL$\downarrow$ & Avg.$\uparrow$ & PPL$\downarrow$ & Avg.$\uparrow$ & PPL$\downarrow$ & Avg.$\uparrow$ \\
\midrule
$0\%$ & Dense   & 14.17 & 49.10 &  9.75 & 49.09 &  7.81 & 57.95 & 13.08 & 48.48 & 5.84 & 63.89 \\
\midrule
\multirow{6}{*}{$50\%$}
 & Wanda             & 32.98 & 42.29 & 23.49 & 39.98 & 12.92 & 49.95 & 24.00 & 41.71 & 9.64 & 55.81 \\
 & SparseGPT         & 28.37 & 43.17 & 18.82 & 42.32 & 12.32 & 50.20 & 20.33 & 41.93 & 9.30 & 57.33 \\
 & Thanos            & 28.65 & 43.88 & 19.70 & 41.62 & 12.26 & 50.81 & 20.85 & 42.10 & 9.34 & 56.82 \\
 & ADMM              & 26.63 & 44.56 & 17.35 & 43.05 & 11.61 & 52.28 & 19.70 & 42.28 & 9.12 & 57.50 \\
 & MaskLLM$^{\dagger}$ & 15.22 & 40.91 & 12.82 & 43.39 & 12.93 & 42.59 & --- & --- & 9.17 & 55.09 \\
 & \method{}         & \textbf{11.89} & \textbf{44.93} & \textbf{11.29} & \textbf{44.81} & \textbf{8.67} & \textbf{54.44} & \textbf{14.09} & \textbf{44.53} & \textbf{7.66} & \textbf{57.71} \\
\midrule
\multirow{5}{*}{$60\%$}
 & Wanda     & 90.50 & 36.64 & 71.53 & 33.83 & 31.13 & 38.77 & 83.42 & 34.66 & 21.66 & 45.19 \\
 & SparseGPT & 60.95 & 38.65 & 47.98 & 36.78 & 22.00 & 43.78 & 40.56 & 37.29 & 15.30 & 49.63 \\
 & Thanos    & 62.22 & 38.94 & 46.78 & 36.95 & 22.48 & 42.49 & 44.29 & 37.64 & 16.10 & 48.87 \\
 & ADMM      & 50.55 & 40.41 & 33.87 & 38.73 & 19.14 & 44.99 & 33.41 & 38.33 & 14.10 & 50.61 \\
 & \method{} & \textbf{13.16} & \textbf{43.41} & \textbf{13.06} & \textbf{42.42} & \textbf{9.77} & \textbf{50.39} & \textbf{15.66} & \textbf{41.53} & \textbf{8.82} & \textbf{54.47} \\
\bottomrule
\end{tabular}
\end{table*}

\paragraph{Ablations.}
\Cref{tab:ablation} ablates the key components of \method{} on Qwen-2.5
0.5B at $50\%$ unstructured sparsity. Removing the weight regularizer
($\lambda_2{=}0$) causes the largest accuracy drop ($44.93 \to 42.87$) and
a $1.67$-point PPL increase, confirming that magnitude-aware stabilization
is the most load-bearing ingredient when weights are frozen. Disabling
either the scale schedule (fixed $\alpha$) or the temperature schedule
(fixed $\tau$) costs roughly $2$ PPL points each while keeping average
accuracy within $0.5$ points of the full method, indicating that the two
schedules contribute primarily to the sharpness of the final mask rather
than to where it lands. Random initialization (no Wanda warm start)
leaves the average zero-shot accuracy essentially unchanged ($44.94$
vs.\ $44.93$) but worsens PPL by $2.54$ points ($14.43$ vs.\ $11.89$),
suggesting that the warm start primarily improves language-modeling
quality rather than zero-shot accuracy at this training budget.
Per-task numbers are in \Cref{app:ablation}.

\begin{table}[t]
\caption{Ablations on Qwen-2.5 0.5B at $50\%$ unstructured sparsity.
Per-task breakdowns are in \Cref{app:ablation}.}
\label{tab:ablation}
\centering
\small
\begin{tabular}{lcc}
\toprule
Variant & PPL $\downarrow$ & Avg.\ Acc.\ $\uparrow$ \\
\midrule
\method{} (full)      & \textbf{11.89} & 44.93 \\
$\lambda_2{=}0$       & 13.56 & 42.87 \\
Random init           & 14.43 & \textbf{44.94} \\
Fixed $\alpha$        & 14.03 & 44.44 \\
Fixed $\tau$          & 13.99 & 44.52 \\
\bottomrule
\end{tabular}
\end{table}

\begin{figure*}[t]
\centering
\includegraphics[width=1.5\columnwidth]{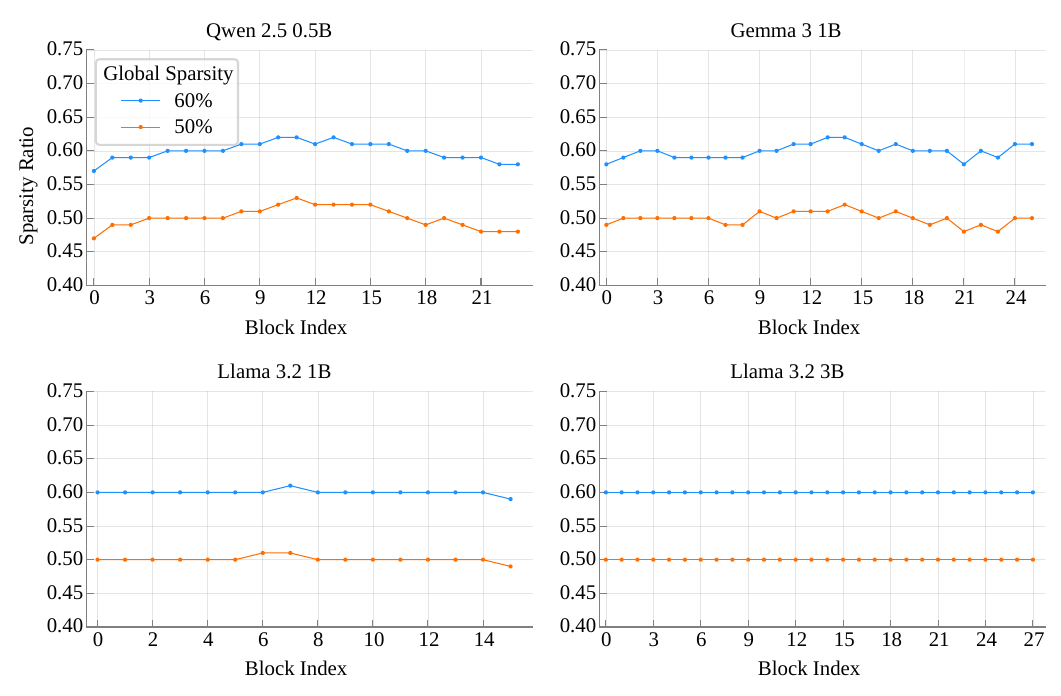}
\caption{Learned per-block density at a global unstructured sparsity
budget of $50\%$ and $60\%$ for Qwen-2.5 0.5B, Gemma-3 1B, LLaMA-3.2 1B, and
LLaMA-3.2 3B. Masks are initialized from Wanda and trained for $2{,}000$
steps. The grey dashed line marks the global budget.}
\label{fig:alloc}
\end{figure*}

\paragraph{Sparsity allocation.}
We study how \method{} distributes sparsity across transformer blocks under
the \emph{global} budget of \Cref{eq:sparsity}. \Cref{fig:alloc} shows the
learned per-block densities. In all four models at both sparsity levels,
\method{} converges to a near-uniform per-block allocation, with only
minor boundary effects at the earliest and latest blocks. This contrasts
with $2{:}4$-constrained methods such as PATCH, which exhibit non-trivial
inter-block variation under the same end-to-end objective.

We read this as a regime-specific observation rather than a general claim
against non-uniform allocation: under end-to-end optimization with a
global budget, at the models and sparsity levels we study, we do not
observe accuracy headroom from non-uniform per-block budgeting. A uniform
per-block allocation is therefore a reasonable default in this regime.
Whether the same holds at higher sparsities, at larger scales, or under
joint weight-mask optimization is an open question.

\paragraph{Runtime and compute.}
\Cref{tab:runtime} reports wall-clock mask-learning time for \method{}.
Models up to $3$B are trained with data parallelism on $4\times$H100;
for LLaMA-3.1 8B we use model parallelism (no data parallelism) on the
same $4\times$H100 node.

\begin{table}[ht]
\caption{\method{} mask-learning time, converted to GPU-hours. The
$0.5$B--$3$B models use data parallelism on $4\times$H100; the $8$B
model uses model parallelism on the same node.}
\label{tab:runtime}
\centering
\small
\begin{tabular}{lrr}
\toprule
Model            & Wall-clock (4$\times$H100) & GPU-hours \\
\midrule
Qwen-2.5 0.5B    & $5$h $40$m                 & $22.7$ \\
LLaMA-3.2 1B     & $7$h $55$m                 & $31.7$ \\
Gemma-3 1B       & $16$h                      & $64$ \\
LLaMA-3.2 3B     & $19$h                      & $76$ \\
LLaMA-3.1 8B     & $69$h                      & $276$ \\
\bottomrule
\end{tabular}
\end{table}

\method{} is more expensive than one-shot layer-wise methods such as
Wanda, SparseGPT, Thanos, and ADMM, which finish in minutes on a single
GPU. In exchange, \method{} delivers the accuracy gains reported above.
Because mask learning is a one-time, offline preprocessing step and the
learned mask is reused at every deployment, we view the additional
compute as a favorable trade-off against the accuracy improvement.

Compared to the other learnable-mask line, \method{} is substantially
cheaper. MaskLLM~\cite{maskllm} reports $1{,}280$ A100 GPU-hours for a
$7$B model and $2{,}304$ A100 GPU-hours for a $13$B model. Per NVIDIA's
published specifications, the H100 delivers higher tensor-core throughput
and memory bandwidth than the A100~\cite{nvidia_h100},
so per-GPU-hour numbers across the two platforms are not directly
comparable; even after generous allowance for the hardware gap,
\method's $276$ GPU-hours on LLaMA-3.1 8B is well below MaskLLM's
reported cost at comparable scale. Two factors in the
\method{} formulation contribute to this gap. First, \method{} trains one
Bernoulli logit per weight, i.e., $1$ mask parameter per weight entry.
MaskLLM parameterizes each $2{:}4$ group of $4$ weights with $6$ logits,
i.e., $1.5$ mask parameters per weight entry, so the trainable state is
$1.5\times$ larger for an equivalently-sized model. Second, a per-weight
Bernoulli is a simpler operation than a per-group softmax over $6$
patterns, which reduces the per-step cost of the mask forward and
backward.

\paragraph{Kernel compatibility.}
\method{} produces masks in the $50\%$--$60\%$ unstructured regime, which
is exactly the regime that recent accelerated kernels target, so the
masks it outputs are deployable without any change to the kernel stack.
We quote the reported numbers from those kernels; we do not measure them
ourselves. Flash-LLM~\cite{flashllm} reports up to $2.9\times$ faster
SpMM than Sputnik and up to $1.5\times$ faster than SparTA, which
translates to up to $3.8\times$ higher end-to-end throughput over
DeepSpeed and $3.6\times$ over FasterTransformer on OPT-30B/66B/175B
under unstructured sparsity. SpInfer~\cite{spinfer} improves on this
line, reporting up to $2.14\times$ faster SpMM than Flash-LLM and
$2.27\times$ faster than SparTA across $30\%$--$70\%$ sparsity, with end-
to-end inference speedups up to $1.58\times$ and surpassing cuBLAS from
as low as $30\%$ sparsity. MACKO~\cite{macko}, designed explicitly for
the low-sparsity regime, reports $1.2$--$1.5\times$ speedup and $1.5\times$
memory reduction over dense fp16 at $50\%$ sparsity, with a $1.5\times$
end-to-end inference speedup on a $50\%$-sparse LLaMA-2 7B. Beyond
commodity GPUs, wafer-scale dataflow accelerators~\cite{cerebras}
report near-linear speedup with unstructured sparsity on GPT-class
models. \method's
$50\%$--$60\%$ unstructured masks feed directly into all of these
kernels; the end-to-end deployment speedup is the composition of
\method's accuracy-preserving mask with the speedups the kernels
deliver.

\section{Discussion and Limitations}
\label{sec:discussion}

\paragraph{Mask learning memory overhead.}
The per-weight logit matrix $P$ has the same shape as $W$ and therefore
roughly doubles the trainable state during mask learning, on top of the
optimizer state $P$ introduces (momentum and second-moment buffers). In
practice, however, this is not the dominant component of GPU memory at
our sequence length. At sequence length $4096$ the activations stored
for backpropagation are the majority of the memory footprint, and the
$P$-plus-optimizer state fits alongside them for every model up to $3$B
on a single H100. Accordingly, we train masks for $0.5$B--$3$B models
with data parallelism across $4\times$H100 within one node: the entire
model, its activations, $P$, and $P$'s optimizer state fit on each GPU,
and the four GPUs process four shards of the calibration batch in
parallel. For LLaMA-3.1 8B, the model plus activations at sequence
length $4096$ no longer fit on a single H100, so we switch to model
parallelism within the same $4\times$H100 node. The $P$-induced memory
doubling can be further reduced by partitioning $P$ and its optimizer
state across data-parallel ranks using ZeRO-style optimizer
sharding~\cite{zero}, or by adopting optimizers that compress
curvature state via low-rank updates~\cite{mkor}; we did not need
either at our scales but they are drop-in extensions for larger models. Note that $P$ is discarded after
mask learning; deployment memory is unaffected and matches a standard
sparse checkpoint.

\paragraph{Frozen weights as a scope choice.}
\method{} keeps $W$ fixed during mask learning. This is not a compute
concession: freezing $W$ preserves pretrained calibration, isolates the
mask as the learned object, and keeps deployment pipelines simple. Joint
weight-and-mask optimization is a natural extension, and can be layered
on top of \method{} as a short fine-tuning stage.

\paragraph{Kernel compatibility, not kernel speedup.}
We do not claim downstream inference speedups as contributions of
\method. \method{} produces unstructured masks in the density regime
already targeted by existing kernels; the reported end-to-end speedup is
the composition of \method's accuracy improvement with the speedups those
kernels deliver.

\section{Conclusion}
\label{sec:conclusion}

We identified a combinatorial obstruction that prevents the
categorical-over-patterns parameterization used by MaskLLM and PATCH from
extending to unstructured sparsity, and we proposed \method, a per-weight
Bernoulli-via-Gumbel-sigmoid reformulation that makes end-to-end
unstructured mask learning tractable. On five LLM families at $50\%$ and
$60\%$ sparsity, \method{} improves the six-task average zero-shot accuracy
over ADMM, the best layer-wise baseline in our sweep, by $+2.59$ points
on average across the ten (model, sparsity) settings and by up to
$+5.40$ points on LLaMA-3.2 1B at $60\%$ sparsity, while
keeping pretrained weights frozen. Because the resulting masks are in the $50\%$--$60\%$
unstructured regime, they are directly consumable by existing accelerated
kernels. We believe the per-weight reformulation opens a practical path to
end-to-end unstructured compression of large foundation models.

\section*{Impact Statement}
\method{} reduces the memory and compute cost of deploying large language
models by producing $50\%$--$60\%$ unstructured masks that are consumable
by existing accelerated kernels. The intended impact is to lower the
barrier to running capable models on commodity hardware and in
resource-constrained settings, and to reduce the energy footprint of
inference at scale. The same compression that enables lower-cost inference
can also make it easier to deploy models outside of supervised
environments, and compressed models may exhibit different failure modes
on long-tail inputs than their dense counterparts; downstream users
applying \method{} to safety-critical deployments should re-evaluate on
the distributions they actually care about rather than relying solely on
our headline benchmarks. Because \method{} keeps pretrained weights frozen
and only trains a mask, it does not introduce new training-data
provenance concerns beyond those already present in the base models.

\newpage

\bibliography{leap}

@inproceedings{spinfer,
  author    = {Ruibo Fan and Xiangrui Yu and Peijie Dong and Zeyu Li and Gu Gong and Qiang Wang and Wei Wang and Xiaowen Chu},
  title     = {{SpInfer}: Leveraging Low-Level Sparsity for Efficient Large Language Model Inference on {GPU}s},
  booktitle = {Proceedings of the Twentieth European Conference on Computer Systems (EuroSys)},
  year      = {2025}
}

@inproceedings{zero,
  author    = {Samyam Rajbhandari and Jeff Rasley and Olatunji Ruwase and Yuxiong He},
  title     = {{ZeRO}: Memory Optimizations Toward Training Trillion Parameter Models},
  booktitle = {Proceedings of the International Conference for High Performance Computing, Networking, Storage and Analysis (SC)},
  year      = {2020}
}

@inproceedings{flashllm,
  author    = {Haojun Xia and Zhen Zheng and Yuchao Li and Donglin Zhuang and Zhongzhu Zhou and Xiafei Qiu and Yong Li and Wei Lin and Shuaiwen Leon Song},
  title     = {{Flash-LLM}: Enabling Cost-Effective and Highly-Efficient Large Generative Model Inference with Unstructured Sparsity},
  booktitle = {Proceedings of the VLDB Endowment (PVLDB), Vol.\ 17, No.\ 2},
  pages     = {211--224},
  year      = {2023}
}

@article{macko,
  author  = {Vladim{\'i}r Macko and Vladim{\'i}r Bo{\v{z}}a},
  title   = {{MACKO}: Sparse Matrix-Vector Multiplication for Low Sparsity},
  journal = {arXiv preprint arXiv:2511.13061},
  year    = {2025}
}

@inproceedings{obs,
  author    = {Babak Hassibi and David G. Stork},
  title     = {Second order derivatives for network pruning: Optimal Brain Surgeon},
  booktitle = {Advances in Neural Information Processing Systems},
  year      = {1993}
}

@inproceedings{wanda,
  author    = {Mingjie Sun and Zhuang Liu and Anna Bair and J. Zico Kolter},
  title     = {A Simple and Effective Pruning Approach for Large Language Models},
  booktitle = {International Conference on Learning Representations (ICLR)},
  year      = {2024}
}

@inproceedings{sparsegpt,
  author    = {Elias Frantar and Dan Alistarh},
  title     = {{SparseGPT}: Massive Language Models Can Be Accurately Pruned in One-Shot},
  booktitle = {International Conference on Machine Learning (ICML)},
  year      = {2023}
}

@article{thanos,
  author  = {Ivan Ilin and Peter Richt{\'a}rik},
  title   = {{Thanos}: A Block-wise Pruning Algorithm for Efficient Large Language Model Compression},
  journal = {arXiv preprint arXiv:2504.05346},
  year    = {2025},
  url     = {https://arxiv.org/abs/2504.05346}
}

@article{admm,
  author  = {Vladim{\'i}r Bo{\v{z}}a},
  title   = {Fast and Effective Weight Update for Pruned Large Language Models},
  journal = {arXiv preprint arXiv:2401.02938},
  year    = {2024}
}

@inproceedings{maskllm,
  author    = {Gongfan Fang and Hongxu Yin and Saurav Muralidharan and Greg Heinrich and Jeff Pool and Jan Kautz and Pavlo Molchanov and Xinchao Wang},
  title     = {{MaskLLM}: Learnable Semi-Structured Sparsity for Large Language Models},
  booktitle = {Advances in Neural Information Processing Systems (NeurIPS)},
  year      = {2024}
}

@article{patch,
  author  = {Younes Hourri and Mohammad Mozaffari and Maryam Mehri Dehnavi},
  title   = {{PATCH}: Learnable Tile-level Hybrid Sparsity for Large Language Models},
  journal = {arXiv preprint arXiv:2509.23410},
  year    = {2025}
}

@inproceedings{elsa,
  author    = {Kwanhee Lee and Hyeondo Jang and Dongyeop Lee and Dan Alistarh and Namhoon Lee},
  title     = {The Unseen Frontier: Pushing the Limits of {LLM} Sparsity with Surrogate-Free {ADMM}},
  booktitle = {International Conference on Learning Representations (ICLR)},
  year      = {2026}
}

@misc{slimpajama,
  author       = {Daria Soboleva and Faisal Al-Khateeb and Robert Myers and Jacob R. Steeves and Joel Hestness and Nolan Dey},
  title        = {{SlimPajama}: A 627B token cleaned and deduplicated version of {RedPajama}},
  howpublished = {\url{https://huggingface.co/datasets/cerebras/SlimPajama-627B}},
  year         = {2023}
}

@article{c4,
  author  = {Colin Raffel and Noam Shazeer and Adam Roberts and Katherine Lee and Sharan Narang and Michael Matena and Yanqi Zhou and Wei Li and Peter J. Liu},
  title   = {Exploring the Limits of Transfer Learning with a Unified Text-to-Text Transformer},
  journal = {Journal of Machine Learning Research},
  volume  = {21},
  number  = {140},
  pages   = {1--67},
  year    = {2020}
}

@inproceedings{piqa,
  author    = {Yonatan Bisk and Rowan Zellers and Ronan Le Bras and Jianfeng Gao and Yejin Choi},
  title     = {{PIQA}: Reasoning about Physical Commonsense in Natural Language},
  booktitle = {AAAI Conference on Artificial Intelligence},
  year      = {2020}
}

@article{arc,
  author  = {Peter Clark and Isaac Cowhey and Oren Etzioni and Tushar Khot and Ashish Sabharwal and Carissa Schoenick and Oyvind Tafjord},
  title   = {Think you have Solved Question Answering? Try {ARC}, the {AI2} Reasoning Challenge},
  journal = {arXiv preprint arXiv:1803.05457},
  year    = {2018}
}

@inproceedings{winogrande,
  author    = {Keisuke Sakaguchi and Ronan Le Bras and Chandra Bhagavatula and Yejin Choi},
  title     = {{WinoGrande}: An Adversarial {Winograd} Schema Challenge at Scale},
  booktitle = {AAAI Conference on Artificial Intelligence},
  year      = {2020}
}

@inproceedings{obqa,
  author    = {Todor Mihaylov and Peter Clark and Tushar Khot and Ashish Sabharwal},
  title     = {Can a Suit of Armor Conduct Electricity? A New Dataset for Open Book Question Answering},
  booktitle = {Empirical Methods in Natural Language Processing (EMNLP)},
  year      = {2018}
}

@inproceedings{mmlu,
  author    = {Dan Hendrycks and Collin Burns and Steven Basart and Andy Zou and Mantas Mazeika and Dawn Song and Jacob Steinhardt},
  title     = {Measuring Massive Multitask Language Understanding},
  booktitle = {International Conference on Learning Representations (ICLR)},
  year      = {2021}
}

@inproceedings{wikitext2,
  author    = {Stephen Merity and Caiming Xiong and James Bradbury and Richard Socher},
  title     = {Pointer Sentinel Mixture Models},
  booktitle = {International Conference on Learning Representations (ICLR)},
  year      = {2017}
}

@article{qwen25,
  author  = {An Yang and Baosong Yang and Beichen Zhang and Binyuan Hui and Bo Zheng and Bowen Yu and Chengyuan Li and Dayiheng Liu and Fei Huang and Haoran Wei and Huan Lin and Jian Yang and Jianhong Tu and Jianwei Zhang and Jianxin Yang and Jiaxi Yang and Jingren Zhou and Junyang Lin and Kai Dang and Keming Lu and Keqin Bao and Kexin Yang and Le Yu and Mei Li and Mingfeng Xue and Pei Zhang and Qin Zhu and Rui Men and Runji Lin and Tianhao Li and Tingyu Xia and Xingzhang Ren and Xuancheng Ren and Yang Fan and Yang Su and Yichang Zhang and Yu Wan and Yuqiong Liu and Zeyu Cui and Zhenru Zhang and Zihan Qiu},
  title   = {{Qwen2.5} Technical Report},
  journal = {arXiv preprint arXiv:2412.15115},
  year    = {2024}
}

@article{llama32,
  author  = {Aaron Grattafiori and Abhimanyu Dubey and Abhinav Jauhri and Abhinav Pandey and Abhishek Kadian and Ahmad Al-Dahle and Aiesha Letman and Akhil Mathur and Alan Schelten and Alex Vaughan and others},
  title   = {The {Llama 3} Herd of Models},
  journal = {arXiv preprint arXiv:2407.21783},
  year    = {2024}
}

@article{gemma3,
  author  = {Gemma Team and Aishwarya Kamath and Johan Ferret and Shreya Pathak and Nino Vieillard and Ramona Merhej and Sarah Perrin and Tatiana Matejovicova and Alexandre Ram{\'e} and Morgane Rivi{\`e}re and others},
  title   = {{Gemma 3} Technical Report},
  journal = {arXiv preprint arXiv:2503.19786},
  year    = {2025}
}

@misc{lmeval,
  author       = {Leo Gao and Jonathan Tow and Baber Abbasi and Stella Biderman and Sid Black and Anthony DiPofi and Charles Foster and Laurence Golding and Jeffrey Hsu and Alain Le Noac'h and Haonan Li and Kyle McDonell and Niklas Muennighoff and Chris Ociepa and Jason Phang and Laria Reynolds and Hailey Schoelkopf and Aviya Skowron and Lintang Sutawika and Eric Tang and Anish Thite and Ben Wang and Kevin Wang and Andy Zou},
  title        = {A framework for few-shot language model evaluation},
  year         = {2024},
  publisher    = {Zenodo},
  version      = {v0.4.3},
  doi          = {10.5281/zenodo.12608602},
  howpublished = {\url{https://github.com/EleutherAI/lm-evaluation-harness}}
}

@inproceedings{l0reg,
  author    = {Christos Louizos and Max Welling and Diederik P. Kingma},
  title     = {Learning Sparse Neural Networks through $L_0$ Regularization},
  booktitle = {International Conference on Learning Representations (ICLR)},
  year      = {2018}
}

@inproceedings{contsparse,
  author    = {Pedro Savarese and Hugo Silva and Michael Maire},
  title     = {Winning the Lottery with Continuous Sparsification},
  booktitle = {Advances in Neural Information Processing Systems (NeurIPS)},
  year      = {2020}
}

@techreport{cerebras,
  author      = {Sean Lie},
  title       = {Harnessing the Power of Sparsity for Large {GPT} {AI} Models},
  institution = {Cerebras Systems},
  year        = {2022}
}

@article{optima,
  author  = {Mohammad Mozaffari and Samuel Kushnir and Maryam Mehri Dehnavi and Amir Yazdanbakhsh},
  title   = {{OPTIMA}: Optimal One-shot Pruning for {LLM}s via Quadratic Programming Reconstruction},
  journal = {arXiv preprint arXiv:2512.13886},
  year    = {2025}
}

@inproceedings{slim,
  author    = {Mohammad Mozaffari and Amir Yazdanbakhsh and Maryam Mehri Dehnavi},
  title     = {{SLiM}: One-shot Quantized Sparse Plus Low-rank Approximation of {LLM}s},
  booktitle = {International Conference on Machine Learning (ICML)},
  year      = {2025}
}

@article{slope,
  author    = {Mohammad Mozaffari and Amir Yazdanbakhsh and Zhao Zhang and Maryam Mehri Dehnavi},
  title     = {{SLoPe}: Double-Pruned Sparse Plus Lazy Low-Rank Adapter Pretraining of {LLM}s},
  journal   = {arXiv preprint arXiv:2405.16325},
  year      = {2024}
}

@inproceedings{mkor,
  author    = {Mohammad Mozaffari and Sikan Li and Zhao Zhang and Maryam Mehri Dehnavi},
  title     = {{MKOR}: Momentum-Enabled Kronecker-Factor-Based Optimizer Using Rank-1 Updates},
  booktitle = {Advances in Neural Information Processing Systems (NeurIPS)},
  year      = {2023}
}

@misc{nvidia_h100,
  author       = {{NVIDIA}},
  title        = {{NVIDIA H100 Tensor Core GPU}},
  howpublished = {\url{https://www.nvidia.com/en-us/data-center/h100/}},
  year         = {2023},
  note         = {Accessed: 2026-04-24}
}
\bibliographystyle{icml2026}

\newpage
\appendix
\onecolumn
\section{Hyperparameters}
\label{app:hparams}

\Cref{tab:hparams} lists the hyperparameters used in all \method{}
experiments in the main paper. Hyperparameters were tuned on the smallest
model (Qwen-2.5 0.5B) and reused for larger models.

\begin{table}[h]
\caption{Hyperparameters used for \method{} experiments.}
\label{tab:hparams}
\centering
\small
\begin{tabular}{ll}
\toprule
Hyperparameter & Value \\
\midrule
Learning rate          & \{$10^{-2}$, $10^{-3}$\} \\
Temperature pair $(\tau_0, \tau_T)$ & $(4.00, 0.05)$ \\
Scale pair $(\alpha_0, \alpha_T)$   & $(25, 350)$ \\
Weight regularizer $\lambda_2$      & $10$ \\
Initial mask strength $s$           & $3$ \\
Weight decay                        & $0$ \\
Batch size                          & $256$ \\
Sequence length                     & $4096$ \\
Mask training steps                 & $2{,}000$ \\
Calibration corpus                  & SlimPajama~\cite{slimpajama} \\
\bottomrule
\end{tabular}
\end{table}

\section{Per-Task Zero-Shot Results}
\label{app:per-task}

This appendix reports the per-task breakdown behind the averaged numbers
in \Cref{tab:summary}. Each table covers one model at $50\%$ and $60\%$
unstructured sparsity across the six zero-shot tasks (MMLU, PIQA, ARC-E,
ARC-C, Winogrande, OBQA) plus WikiText2 perplexity at sequence length
$4096$. The MaskLLM$^{\dagger}$ rows are $2{:}4$ semi-structured at
$50\%$ density and are reproduced from~\cite{patch}.

\begin{table*}[h]
\caption{Per-task results on Qwen-2.5 0.5B. PPL is on WikiText2 (lower
is better). Other columns are zero-shot accuracy in percent.}
\label{tab:qwen}
\centering
\small
\setlength{\tabcolsep}{3pt}
\begin{tabular}{llccccccccc}
\toprule
Method & Sparsity & PPL $\downarrow$ & MMLU & PIQA & ARC-E & ARC-C & Wino. & OBQA & Avg. \\
\midrule
Dense     & 0\%  & 14.17 & 24.95 & 74.81 & 71.93 & 35.41 & 58.72 & 28.80 & 49.10 \\
\midrule
Wanda             & \multirow{6}{*}{50\%} & 32.98 & 22.92 & 67.46 & 60.77 & 26.37 & 56.04 & 20.20 & 42.29 \\
SparseGPT         &                       & 28.37 & 24.81 & 69.10 & 61.15 & 27.13 & 55.64 & 21.20 & 43.17 \\
Thanos            &                       & 28.65 & 23.09 & 69.75 & 62.16 & 27.99 & 56.51 & 23.80 & 43.88 \\
ADMM              &                       & 26.63 & 24.05 & 70.08 & 63.80 & 27.73 & 56.51 & 25.20 & 44.56 \\
MaskLLM$^{\dagger}$ &                     & 15.22 & 25.11 & 67.03 & 56.57 & 23.98 & 52.57 & 20.20 & 40.91 \\
\method{}         &                       & \textbf{11.89} & 23.80 & \textbf{71.27} & 63.13 & \textbf{27.90} & \textbf{60.03} & 23.20 & \textbf{44.93} \\
\midrule
Wanda     & \multirow{5}{*}{60\%} & 90.50 & 23.02 & 62.13 & 49.87 & 18.34 & 51.07 & 15.40 & 36.64 \\
SparseGPT &                       & 60.95 & 24.54 & 65.61 & 52.02 & 21.59 & 51.54 & 16.60 & 38.65 \\
Thanos    &                       & 62.22 & 24.62 & 64.53 & 52.86 & 20.65 & 52.17 & 18.80 & 38.94 \\
ADMM      &                       & 50.55 & 25.16 & 65.29 & 55.89 & 22.69 & 53.82 & 19.60 & 40.41 \\
\method{} &                       & \textbf{13.16} & 24.44 & \textbf{68.66} & \textbf{60.61} & \textbf{25.09} & \textbf{58.64} & \textbf{23.00} & \textbf{43.41} \\
\bottomrule
\end{tabular}
\end{table*}

\begin{table*}[h]
\caption{Per-task results on Gemma-3 1B.}
\label{tab:gemma}
\centering
\small
\setlength{\tabcolsep}{3pt}
\begin{tabular}{llccccccccc}
\toprule
Method & Sparsity & PPL $\downarrow$ & MMLU & PIQA & ARC-E & ARC-C & Wino. & OBQA & Avg. \\
\midrule
Dense     & 0\%  & 9.75 & 36.92 & 74.27 & 65.53 & 31.31 & 60.30 & 26.20 & 49.09 \\
\midrule
Wanda             & \multirow{6}{*}{50\%} & 23.49 & 26.27 & 64.96 & 51.98 & 23.55 & 54.93 & 18.20 & 39.98 \\
SparseGPT         &                       & 18.82 & 25.64 & 67.85 & 54.55 & 26.53 & 57.14 & 22.20 & 42.32 \\
Thanos            &                       & 19.70 & 25.37 & 67.63 & 52.99 & 27.13 & 54.38 & 22.20 & 41.62 \\
ADMM              &                       & 17.35 & \textbf{27.92} & 69.75 & 56.06 & 26.37 & 56.04 & 22.20 & 43.05 \\
MaskLLM$^{\dagger}$ &                     & 12.82 & 25.03 & 69.91 & 60.27 & 27.65 & 56.27 & 21.20 & 43.39 \\
\method{}         &                       & \textbf{11.29} & 27.39 & \textbf{71.98} & \textbf{60.90} & \textbf{28.24} & \textbf{57.53} & \textbf{22.80} & \textbf{44.81} \\
\midrule
Wanda     & \multirow{5}{*}{60\%} & 71.53 & 22.93 & 59.36 & 39.65 & 18.86 & 50.20 & 12.00 & 33.83 \\
SparseGPT &                       & 47.98 & 23.06 & 62.24 & 43.77 & 21.76 & 52.25 & 17.60 & 36.78 \\
Thanos    &                       & 46.78 & 23.25 & 62.57 & 44.49 & 51.59 & 53.20 & 16.60 & 36.95 \\
ADMM      &                       & 33.87 & \textbf{25.77} & 64.15 & 47.22 & 22.44 & 54.62 & 18.20 & 38.73 \\
\method{} &                       & \textbf{13.06} & 24.90 & \textbf{69.70} & \textbf{57.53} & \textbf{25.60} & \textbf{55.80} & \textbf{21.00} & \textbf{42.42} \\
\bottomrule
\end{tabular}
\end{table*}

\begin{table*}[h]
\caption{Per-task results on LLaMA-3.2 1B.}
\label{tab:llama1b}
\centering
\small
\setlength{\tabcolsep}{3pt}
\begin{tabular}{llccccccccc}
\toprule
Method & Sparsity & PPL $\downarrow$ & MMLU & PIQA & ARC-E & ARC-C & Wino. & OBQA & Avg. \\
\midrule
Dense     & 0\%  & 7.81 & 54.13 & 76.55 & 74.28 & 42.75 & 69.38 & 30.60 & 57.95 \\
\midrule
Wanda             & \multirow{6}{*}{50\%} & 12.92 & 40.79 & 72.03 & 65.45 & 32.34 & 63.69 & 25.40 & 49.95 \\
SparseGPT         &                       & 12.32 & 37.96 & 73.45 & 65.19 & 33.02 & 66.38 & 25.20 & 50.20 \\
Thanos            &                       & 12.26 & 40.11 & 72.80 & 64.77 & 32.85 & 67.72 & 26.60 & 50.81 \\
ADMM              &                       & 11.61 & 42.90 & 74.48 & 66.62 & 34.56 & 66.93 & 28.20 & 52.28 \\
MaskLLM$^{\dagger}$ &                     & 12.93 & 26.28 & 69.10 & 57.41 & 25.85 & 55.48 & 21.40 & 42.59 \\
\method{}         &                       & \textbf{8.67}  & \textbf{44.55} & \textbf{75.19} & \textbf{71.09} & \textbf{39.33} & \textbf{67.48} & \textbf{29.00} & \textbf{54.44} \\
\midrule
Wanda     & \multirow{5}{*}{60\%} & 31.13 & 25.53 & 65.23 & 47.90 & 22.70 & 55.25 & 16.00 & 38.77 \\
SparseGPT &                       & 22.00 & 31.27 & 69.37 & 53.66 & 26.02 & 61.33 & 21.00 & 43.78 \\
Thanos    &                       & 22.48 & 29.23 & 67.63 & 55.01 & 26.02 & 57.85 & 19.20 & 42.49 \\
ADMM      &                       & 19.14 & 33.46 & 69.15 & 57.70 & 27.39 & 59.82 & 22.40 & 44.99 \\
\method{} &                       & \textbf{9.77}  & \textbf{37.55} & \textbf{74.32} & \textbf{66.50} & \textbf{34.81} & \textbf{63.14} & \textbf{26.00} & \textbf{50.39} \\
\bottomrule
\end{tabular}
\end{table*}

\begin{table*}[h]
\caption{Per-task results on LLaMA-3.2 3B.}
\label{tab:llama3b}
\centering
\small
\setlength{\tabcolsep}{3pt}
\begin{tabular}{llccccccccc}
\toprule
Method & Sparsity & PPL $\downarrow$ & MMLU & PIQA & ARC-E & ARC-C & Wino. & OBQA & Avg. \\
\midrule
Dense     & 0\%  & 13.08 & 47.36 & 69.97 & 64.18 & 29.18 & 55.80 & 24.40 & 48.48 \\
\midrule
Wanda     & \multirow{5}{*}{50\%} & 24.00 & 30.52 & 64.09 & 57.41 & 24.06 & 54.38 & 19.80 & 41.71 \\
SparseGPT &                       & 20.33 & 29.38 & 64.74 & 56.52 & 24.15 & 56.20 & 20.60 & 41.93 \\
Thanos    &                       & 20.85 & 28.94 & 65.40 & 55.93 & 24.40 & 56.35 & \textbf{21.60} & 42.10 \\
ADMM      &                       & 19.70 & 29.38 & 65.67 & 55.89 & 25.26 & 56.27 & 21.20 & 42.28 \\
\method{} &                       & \textbf{14.09} & \textbf{34.37} & \textbf{68.44} & \textbf{61.99} & \textbf{26.11} & 55.25 & 21.00 & \textbf{44.53} \\
\midrule
Wanda     & \multirow{5}{*}{60\%} & 83.42 & 23.02 & 59.96 & 43.81 & 18.09 & 50.28 & 12.80 & 34.66 \\
SparseGPT &                       & 40.56 & 22.90 & 61.59 & 48.40 & 21.25 & 52.80 & 16.80 & 37.29 \\
Thanos    &                       & 44.29 & 23.78 & 62.02 & 48.65 & 21.33 & 52.25 & 17.80 & 37.64 \\
ADMM      &                       & 33.41 & 24.22 & 62.40 & 50.13 & 22.44 & 52.96 & 17.80 & 38.33 \\
\method{} &                       & \textbf{15.66} & \textbf{24.66} & \textbf{67.90} & \textbf{56.69} & \textbf{25.09} & \textbf{55.49} & \textbf{19.40} & \textbf{41.53} \\
\bottomrule
\end{tabular}
\end{table*}

\section{Per-Task Ablation Results}
\label{app:ablation}

\Cref{tab:ablation-full} reports the per-task zero-shot breakdown for the
ablation study summarized in \Cref{tab:ablation} (Qwen-2.5 0.5B at $50\%$
unstructured sparsity).

\begin{table*}[h]
\caption{Per-task ablation results on Qwen-2.5 0.5B at $50\%$
unstructured sparsity.}
\label{tab:ablation-full}
\centering
\small
\setlength{\tabcolsep}{4pt}
\begin{tabular}{lcccccccc}
\toprule
Variant & PPL $\downarrow$ & MMLU & PIQA & ARC-E & ARC-C & Wino. & OBQA & Avg. \\
\midrule
\method{} (full)   & \textbf{11.89} & 23.80          & \textbf{71.27} & \textbf{63.13} & \textbf{27.90} & \textbf{60.03} & \textbf{23.20} & 44.93 \\
$\lambda_2{=}0$    & 13.56          & 26.32          & 68.61          & 60.82          & 26.20          & 54.07          & 21.20          & 42.87 \\
Random init        & 14.43          & \textbf{34.16} & 68.44          & 62.84          & 26.96          & 55.41          & 21.80          & \textbf{44.94} \\
Fixed $\alpha$     & 14.03          & 32.31          & 68.34          & 62.96          & 27.22          & 54.38          & 21.40          & 44.44 \\
Fixed $\tau$       & 13.99          & 33.81          & 68.34          & 62.21          & 26.62          & 54.14          & 22.00          & 44.52 \\
\bottomrule
\end{tabular}
\end{table*}

\section{Additional Notes on the Combinatorial Argument}
\label{app:combinatorics}

The obstruction argument in \Cref{sec:method:obstruction} relies only on
the statement that a categorical parameterization over $\mathcal{P}_n$
requires $|\mathcal{P}_n|$ logits. For $\rho$-sparsity at row width $n$,
\[
|\mathcal{P}_n| \;=\; \binom{n}{\rho n},
\]
and Stirling gives
\[
\log_2 \binom{n}{\rho n}
\;=\; n\,H_2(\rho) + o(n),\qquad
H_2(\rho) = -\rho\log_2\rho - (1{-}\rho)\log_2(1{-}\rho).
\]
At $\rho{=}0.5$, $H_2(\rho){=}1$, so the logit table for a single row of
width $n{=}4096$ already requires $\sim 2^{4096}$ entries. No sparsification
of the logit table preserves the unstructured constraint: any coarser
grouping reimposes structural constraints on $\mathcal{P}$. This is why
we argue the per-weight Bernoulli parameterization is the natural
reformulation, not a heuristic alternative.

\end{document}